\documentclass{optica-article}

\journal{opticajournal} 

\articletype{Research Article}

\usepackage{lineno}
\usepackage{chemformula} 
\usepackage[T1]{fontenc} 
\usepackage{siunitx}
\usepackage{booktabs}
\usepackage{multirow}
\usepackage{subfig}
\usepackage[]{graphicx}
\usepackage{MnSymbol,wasysym}
\usepackage{soul,xcolor}

\begin{document}

\title{Optimal Data Generation in Multi-Dimensional Parameter Spaces, using Bayesian Optimization}

\author{M. R. Mahani,\authormark{1*}, Igor A. Nechepurenko,\authormark{1}, Yasmin Rahimof,\authormark{1}, Andreas Wicht,\authormark{1}}

\address{\authormark{1}Ferdinand-Braun-Institut (FBH), Leibniz-Institut f\"{u}r H\"{o}chstfrequenztechnik, Gustav-Kirchhoff-Straße 4, 12489, Berlin, Germany}

\email{\authormark{*}Reza.Mahani@FBH-Berlin.de}

\begin{abstract*} 
Acquiring a substantial number of data points for training accurate machine learning (ML) models is a big challenge in scientific fields where data collection is resource-intensive. Here, we propose a novel approach for constructing a minimal yet highly informative database for training ML models in complex multi-dimensional parameter spaces. To achieve this, we mimic the underlying relation between the output and input parameters using Gaussian process regression (GPR). Using a set of known data, GPR provides predictive means and standard deviation for the unknown data. Given the predicted standard deviation by GPR, we select data points using Bayesian optimization to obtain an efficient database for training ML models. We compare the performance of ML models trained on databases obtained through this method, with databases obtained using traditional approaches. Our results demonstrate that the ML models trained on the database obtained using Bayesian optimization approach consistently outperform the other two databases, achieving high accuracy with a significantly smaller number of data points. Our work contributes to the resource-efficient collection of data in high-dimensional complex parameter spaces, to achieve high precision machine learning predictions.
\end{abstract*}

\section{Introduction}
In certain domains of scientific studies, the challenge lies in unraveling patterns and extracting information from extensive databases \cite{venketeswaran2022recent, thiyagalingam2022scientific}. Contrasting to these areas, there exist various scientific fields where the acquisition of data involves time-consuming experiments or simulations rather than sifting through large data. In these fields, the endeavor to collect large datasets for training accurate machine learning (ML) models in regression problems is often resource-intensive \cite{yao2019intelligent, nielsen2015neural}. A substantial number of data points is always required for efficient training of ML models and consequently their accurate prediction but it is frequently a limiting factor, in terms of time \cite{nadell2019deep, ma2021deep, hammond2019designing, dey2023demonstration, hegde2020deep, hughes2018training}. Thus, the task of generating a minimal number of data points that encapsulate the maximum information content is a paramount objective in data-driven research in these fields.

In pursuit of highly informative yet minimal database for training various ML models, we harness the power of Bayesian optimization (BO) \text{--} a methodology traditionally applied to optimization problems \cite{garnett2023bayesian, shahriari2015taking}. BO has emerged as a powerful tool, leveraging probabilistic modeling to intelligently guide sampling of an unknown multivariate function to find its extremum with minimum number of selected points. This approach can be adjusted to emulate the relation between the input and output parameters (the unknown multivariate function). Using this surrogate model, one can find points within the parameter space for which this function has highest uncertainty to predict their output. A database built using these data contains minimum required points to capture most variations in the data.

We provide an example by building a database that includes the characteristics of Bragg gratings and their respective reflectance spectra. The database is generated using finite-difference time-domain (FDTD) simulations \cite{burr2005balancing, mahani2023data, teixeira2023finite}. A ML model trained on this small database should significantly outperform the same model trained on the same number of data points collected by traditional means (uniform or randomly distributed data points). This challenge is particularly pertinent in applications such as inverse design of optical devices, where achieving high predictive accuracy is of great significance. As an example, designing Bragg gratings to achieve a precise optical response is crucial for fabricating diode lasers with specific applications, ranging from telecommunications to sensors, optical atomic clocks and quantum technologies \cite{lezius2016space,becker2018space,shemshad2012review,lin2022improvement,jin2018high}. Inverse design of these Bragg gratings using ML models \cite{mahani2023data, mahani2023designing} necessitates collecting a substantial amount of data in multi-dimensional parameter spaces to achieve high predictive accuracy. This, in turn, can be time-consuming and resource-intensive. 

In the following, we introduce the methodology, and the steps required for constructing informative database. We briefly introduce the mechanism of BO and present the results. Although this approach is applied to photonics here, the methodology can be easily applied to building any database within the domain of regression analysis in various fields. Showcasing this example will stimulate further exploration of BO in the field of data-driven inverse design and contribute to more efficient and accurate predictions.

\section{Methodology}
Our approach in constructing a minimal yet efficient database for training ML models, is to identify the most informative data points one by one using BO.
In addition to this database we obtain two other ones, using uniform and random distribution of data points. 
These two choices are the most common approaches to generate data when there is limited or no prior knowledge about the data.
Then we can compare the performance of ML models trained on these three databases. The databases include the Bragg grating parameters and the reflectance spectra simulated by varying Bragg gratings' characteristics. We train ML models on these three databases and make prediction on the new data. We then use the coefficient of determination ($R^2$) and the mean-squared-error (MSE) to compare the ML predictive performance.   

\subsection{Constructing Database}
The database is made of six input parameters (length of Bragg grating, depth, width and refractive index of gratings' grooves, chirp value and the order of the grating) that are the characteristics of Bragg gratings, making up a 6D parameter space. The output is the reflectance spectra obtained using 2D FDTD simulations in ANSYS Lumerical. Since the reflectance spectra contains many data points, we fit the top $2/3$ of the main lobe of the reflectance spectra using a Gaussian function with three parameters \cite{mahani2023designing, nechepurenko2023finite}. These three fit parameters are then used as output and describe main characteristics of the Bragg resonance: amplitude, bandwidth and frequency \cite{agrawal2013semiconductor,coldren2012diode}. Therefore, each data point in the database contains six inputs and three output parameters. 

\begin{figure}[htbp]
\includegraphics[height=3.in]{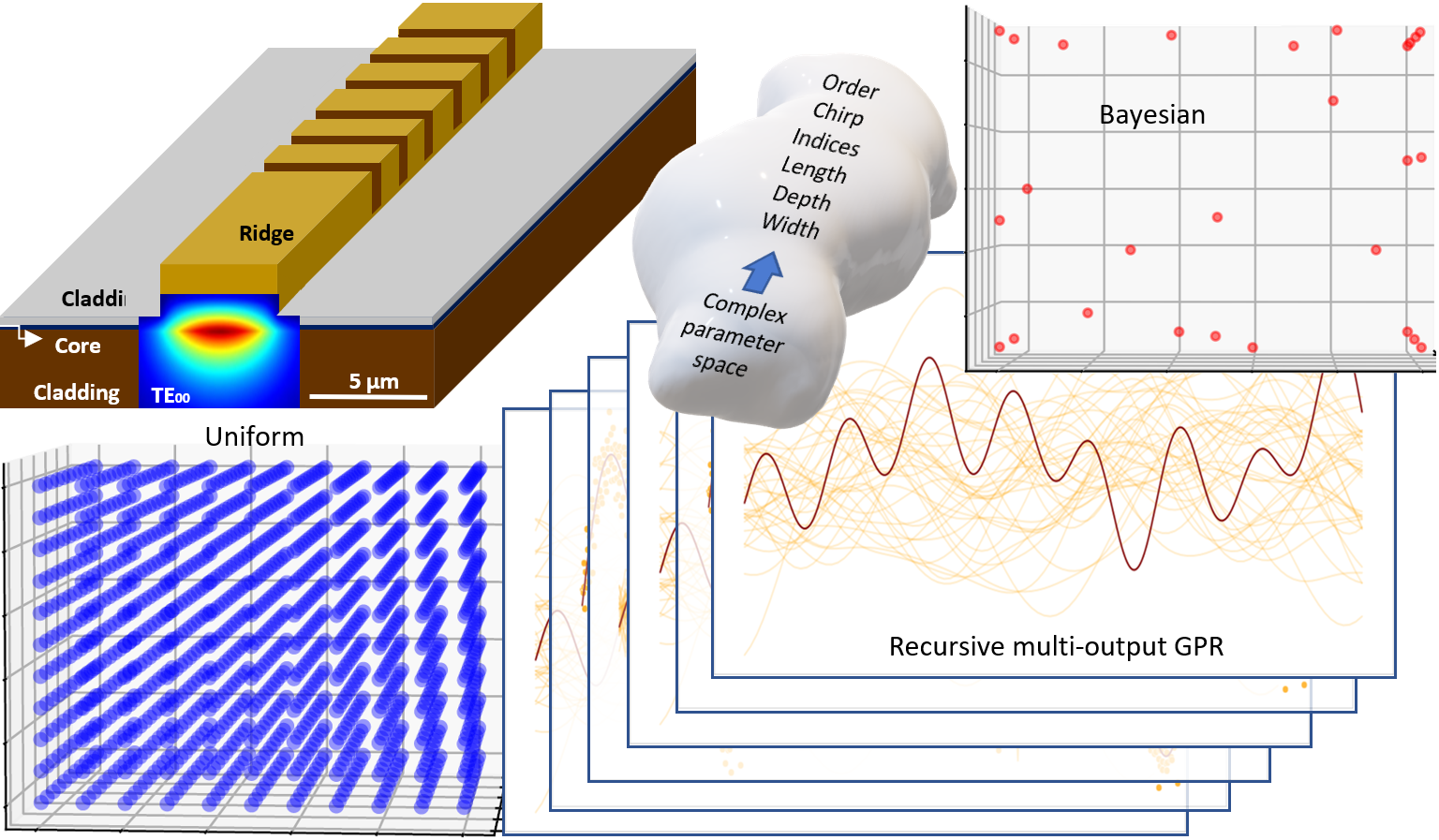}
\caption{(Colour online) The schematic of the structure, parameter space and data acquisition. Through Bayesian optimization approach using multi-output GPR, one can obtain small and more informative data points.}
\label{fig:fig1}
\end{figure}

We build the first database, uniform-based dataset (UBD), using uniform distribution of data points in the parameter space. As an example, for a database in 6D parameter space with 64 data points, we sample each input parameter set with two values equally separated from one another ($2^6=64$). The points are chosen within the technologically relevant range for each parameter set. We call the number of data points that are of the type of $n^6$ (eg, $2^6, 3^6, ...$) in 6D parameter space, evenly divisible. Intermediate points refer to number of points between the evenly divisible points (eg., any number between $2^6=64$ and $3^6=729$). For intermediate points within this UBD, we sample one or more parameter sets with an additional point, spaced uniformly, till we reach the required number.

We calculate three output values for each data point using 2D FDTD simulations Lumerical, as explained before. We have also done 3D FDTD simulations, but due to the time-constraint, these simulations are limited to a 4D parameter space. The time required to obtain a single data point using 3D FDTD is 1272 core-hour, while it takes only 1.36 core-hour for the 2D FDTD. This means generating a uniform database with only three points per each parameter set in 6D parameter space would require 927288 and 1730 core-hour for 3D and 2D FDTD, respectively. To illustrate the results in larger dimension of parameter space with higher number of points, we present the 2D FDTD results. However the 3D FDTD results agree well with 2D FDTD for all the conclusions obtained within this study.

If instead of uniform distribution, the points were chosen randomly, the performance of ML model trained on this database would vary a lot from a random selection to another one. This random selection has to be repeated many times then averaged over all the trials for a reproducible performance. However this is computationally too expensive, due to all the repetitions. Instead of this random database, we combine the uniform selection with random distribution of additional points, uniform-random based dataset (URBD). This means for the intermediate points (clarified above), we take the uniform distribution for the evenly divisible numbers and we add additional points using a random distribution. The number of additional points is divided equally among all the parameter sets (if possible). In the example above, for a database with 64 till 729 data points, we uniformly choose 64 points (UBD), then randomly choose the remaining data points, drawn equally from each parameter set. 

The Bayesian-based dataset (BBD) is built using BO approach. The construction of the database proceeds iteratively, with the following steps:
(a) We randomly choose initial data points (in this case only two data points) in the entire parameter space and obtain the output as explained above.
(b) We train a multi-output GPR model on the existing data points.
(c) We generate a dense uniform input mesh on the entire parameter space. This means we sample each input parameters set with 11 uniformly spaced points.
(d) The trained GPR on the number of data makes a prediction on the dense mesh. This means that GPR trained on the obtained data points provides the predictive means and standard deviation for all the unknown data points on a much finer grid (dense mesh).
(e) We calculate the acquisition function (in this case only the predicted standard deviation) value for all unexplored points in the parameter space. 
(f) We select the data point with the maximum acquisition function value (highest standard deviation) as a newly acquired data point.
(g) We run FDTD simulations for this data point (each data point includes six input parameters of Bragg gratings) to obtain the reflectance spectra and thus output parameters.
(h) This data point with six input and three output parameters is then added to the initial data point in the database.
(i) We repeat steps b-h until we reach the desired number of data points.
Figure \ref{fig:fig1} shows the schematic of the structure, parameter space and data acquisition procedure.

\subsection{Bayesian Optimization}
Bayesian optimization is a powerful and versatile method for optimizing expensive multivariate nonlinear functions. Here, we employ Bayesian optimization to guide the selection of data points. 
The core idea behind this is to model the unknown multivariate function (that connects the Bragg grating's characteristics, input parameters, to their reflectance spectra, output parameters) using a probabilistic and cheap-to-evaluate surrogate model. We use GPR as the surrogate model due to its ability to provide predictive averages and standard deviations (uncertainties) for unobserved points. This predictive measures can then be used to guide the selection of the next data point.

A Gaussian process is defined by a mean function (typically zero) and a covariance or kernel function (we use radial basis function).
Mean function represents the expected value of the function at each input point. Covariance function quantifies the similarity between known and unknown points, expressing the model's confidence in its predictions and provide a predictive standard deviation. As new data points are obtained, the Gaussian process is updated, which leads to updated predictive means and standard deviations\cite{rasmussen2006gaussian}. This probabilistic nature makes GPR well-suited for applications such as Bayesian optimization, where uncertainty plays a crucial role in guiding the search for optimal solutions.

In BO, an acquisition function is used to decide where to sample next, utilizing the Gaussian process's predictive mean and standard deviation. The acquisition function, denoted as $\alpha(x)$, guides the selection of new data points by balancing exploration (sampling uncertain regions) and exploitation (sampling promising regions). It is defined as $\alpha(x)$=$\mu(x)$+$\kappa$$\sigma(x)$, where $\mu(x)$ and $\sigma(x)$ are the predicted mean and standard deviation by the surrogate model at point $x$, respectively. The tunable parameter, $\kappa$, controls the trade-off between exploration and exploitation. In our case, we are only interested in exploring the parameter space, hence, considering only the standard deviation term in the acquisition function ($\kappa=0$). This leads to a more diversified selection of data points, enhancing the ability to capture variations in the data. Once the data points are collected using BO, we can compare the prediction accuracy of ML models trained on this database (BBD) vs other two databases (UBD and URBD). 

\section{Results and Discussions}

To evaluate the effectiveness of each database, we train ML models on one of the databases, then test it on the other database. The data points on the other database are new data to the already trained ML models. We train support vector regression (SVR) and optimized extreme Gradient Boosting (XGBoost) models and assess their performance by means of $R^2$ and MSE. The reason we choose these models because SVR represents a shallow ML model that perform reasonably well with small data but lack the expressive power to capture complex patterns with increasing data. On the other hand, XGBoost represents a complicated algorithm which has proven to be flexible in learning intricate relationships \cite{chen2016xgboost, mahani2023data}.
As we will see in the following not only the complexity of the database, but also the complexity of the ML algorithm plays a crucial role in the overall performance. We always use the same number of data points in each database for training and testing each ML model, to remove any potential bias.

The coefficient of determination, $R^2=1-[{\sum_i (y_{true}-y_{pred})^2}/{\sum_i (y_{true}-y_{mean})^2}]$, evaluates the accuracy of the trained ML models. It measures the proportion of the variance in the output predicted by the model. The mean squared error, $\text{MSE} = \frac{1}{n} \sum_{i=1}^{n} (y_{true}-y_{pred})^2$, measures the average squared difference between the actual values and the predicted ones.  Variables $y_{true}$ and $y_{pred}$ are the actual values of the target feature and the predicted values, respectively.  

\begin{figure}[htbp]
\subfloat[]{\includegraphics[height=1.7in]{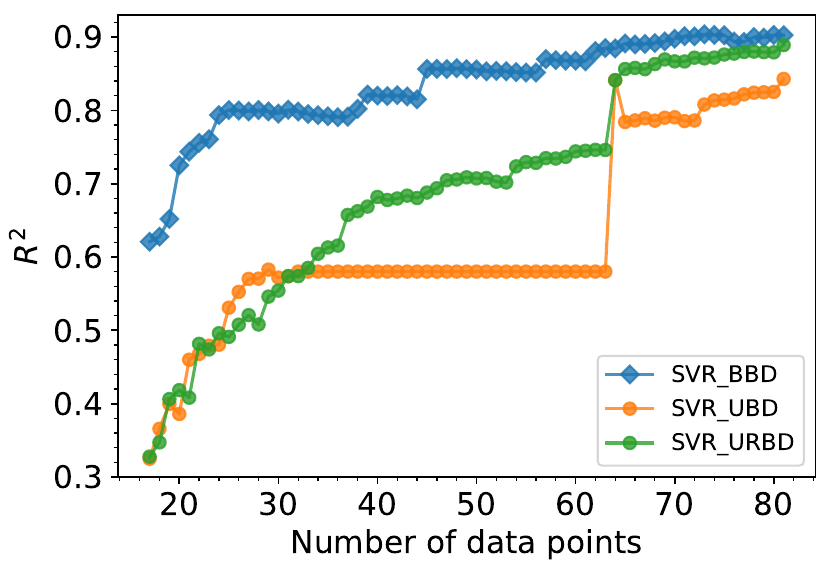}}
\subfloat[]{\includegraphics[height=1.7in]{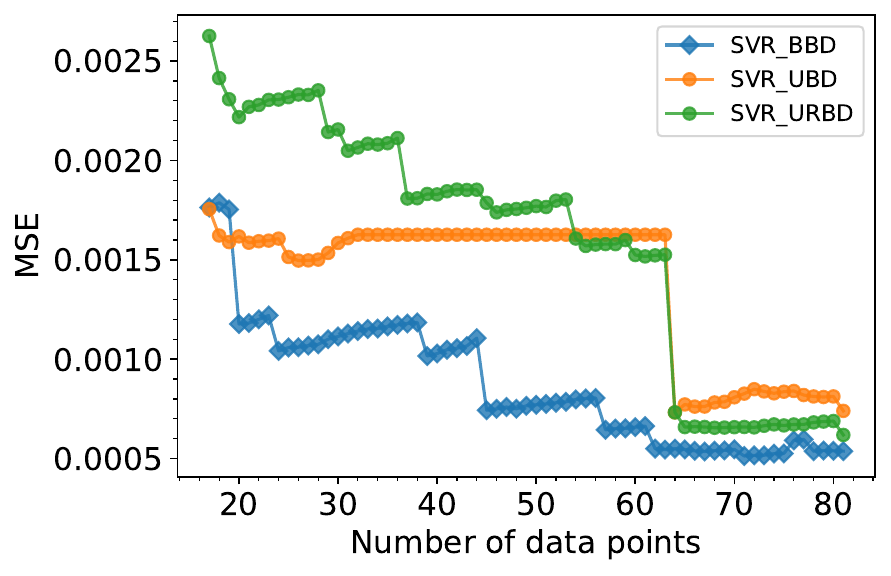}}
\caption{(Colour online) Comparison between the performance of an SVR model trained on three databases (BBD, UBD, URBD). (a) Prediction accuracy, $R^2$, of SVR as a function of number of data points used for training, (b) Prediction mean-squared-error of SVR as a function of number of data points used for training.}
\label{fig:comp1}
\end{figure}

Figure \ref{fig:comp1} shows the comparison between the performance of an SVR model trained on three databases (BBD, UBD, URBD). The performance measures, $R^2$ and MSE, are plotted against the number of data points used for training the SVR model (figures.\ref{fig:comp1}(a) and \ref{fig:comp1}(b) respectively). We have plotted the number of points till an accuracy of $R^2>0.9$ is achieved. We could see from this figure that the SVR model trained on URBD (a combination of uniform and random) outperforms the training on UBD (the uniform database). The gap between the two decreases at evenly divisible points (in this case $2^6$). The SVR trained on BBD clearly outperforms the other two model, however, the gap between them decreases as the number of points increases. This is due to the fact that, SVR is a shallow model that is not best suited for capturing complex relationships in the data. It can reach a performance ceiling more quickly, especially when the dataset becomes sufficiently complex (at higher number of points). As we will see in the following, a more flexible model like XGBoost can capture complex relationships allowing the model to continue benefiting from additional data adding new information.

\begin{figure}[htbp]
\subfloat[]{\includegraphics[height=1.7in]{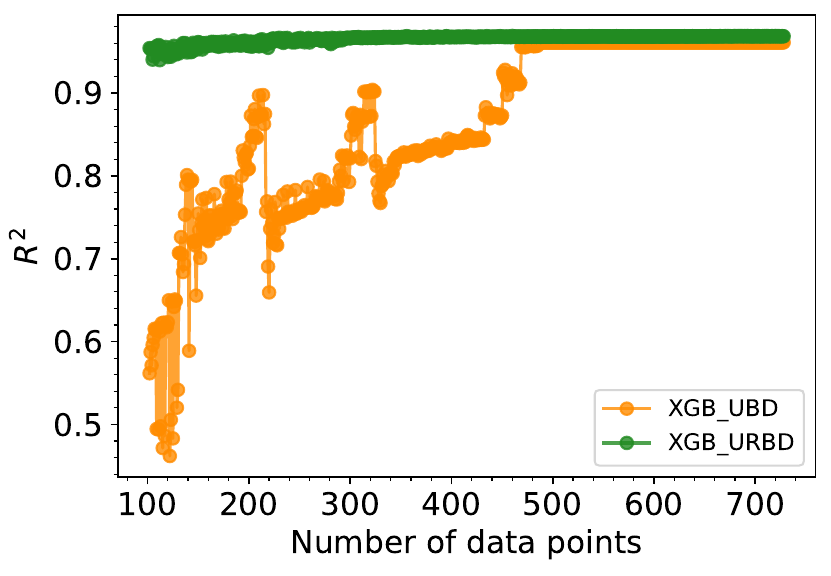}}
\subfloat[]{\includegraphics[height=1.7in]{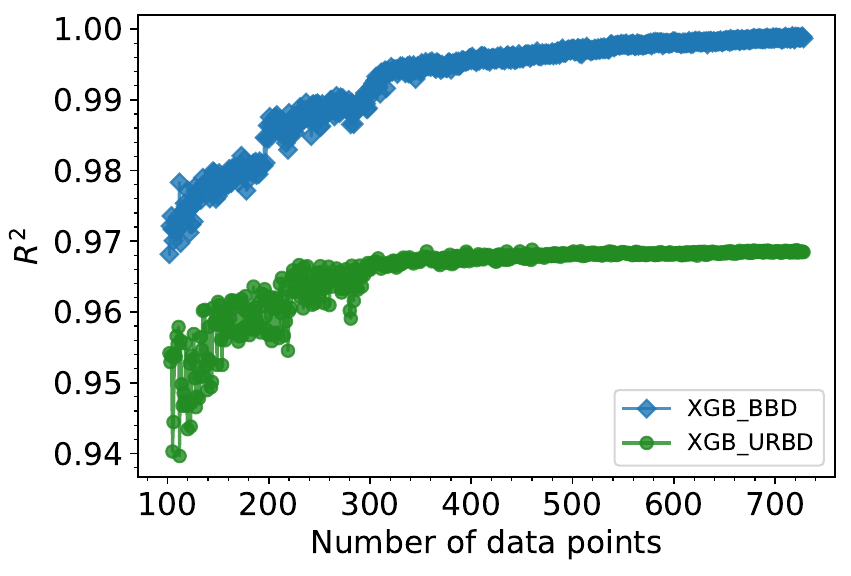}}

\subfloat[]{\includegraphics[height=1.7in]{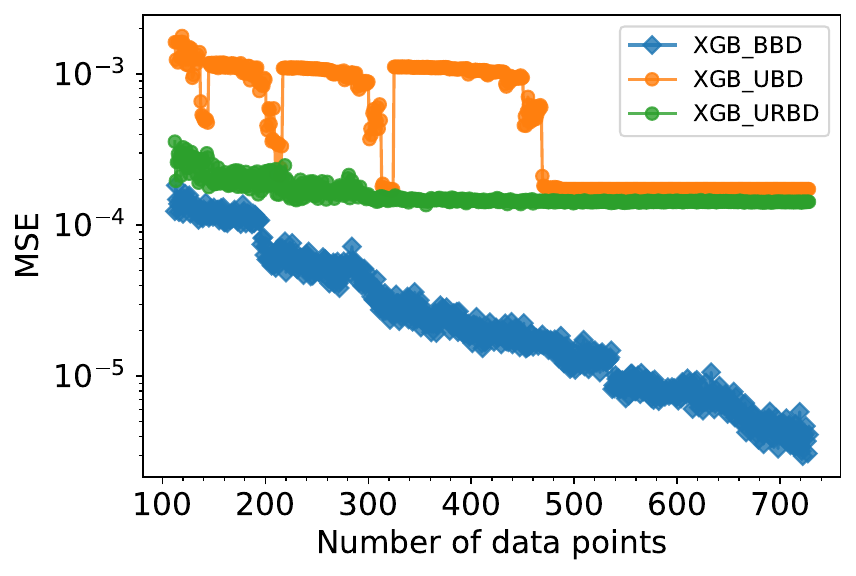}}
\subfloat[]{\includegraphics[height=1.7in]{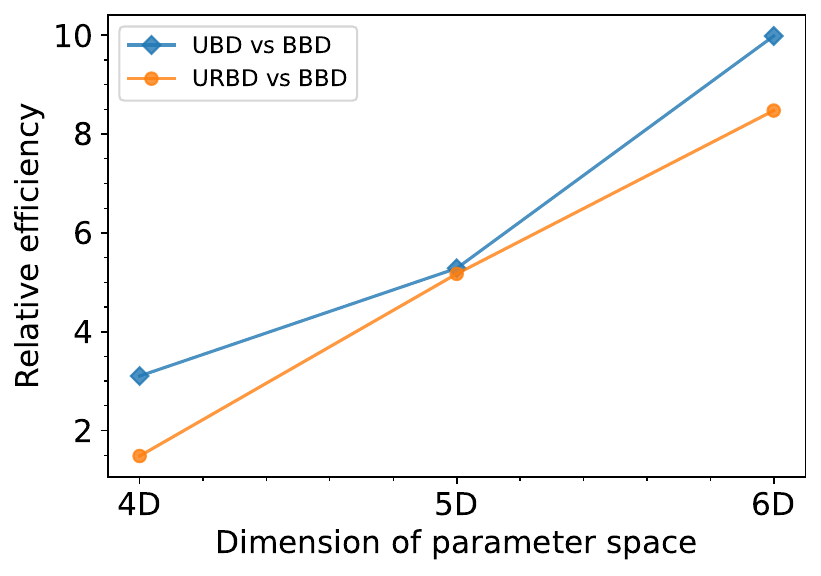}}
\caption{(Colour online) Performance comparison of XGBoost model trained on three different databases (BBD, UBD, URBD). (a) Prediction accuracy, $R^2$, of XGBoost trained on UBD vs URBD as a function of number of data points used for training. (b) Comparison between BBD and URBD. (c) Prediction mean-squared-error of XGBoost trained on BBD, UBD and URBD as a function of number of data points used for training. (d) Relative efficiency as a function of dimension of parameter space. Number of data points required for XGBoost trained on UBD (blue) or URBD (orange) to reach $R^2 \approx 0.97$ divided by the number of data points for BBD to surpass this accuracy. In the legends, XGB is an abbreviation for XGBoost.}
\label{fig:comp2}
\end{figure}

We noticed that $R^2$ sometimes plateaus or fluctuates while adding extra points to the database. This happens regardless of which approach we use for selecting data points. This is a common behaviour for the ML models, where adding single data points to the database does not necessarily always improve $R^2$ \cite{nielsen2015neural}.
If the new data introduces features that are redundant or highly correlated with existing features, or if the distribution of the new data differs significantly from the existing data, the model may struggle to adapt, leading to fluctuations in the performance. Generally, the model initially experience rapid improvements in $R^2$, but as it approaches its capacity to model the data, the gains will slow down.

Figure \ref{fig:comp2} shows the performance comparison of XGBoost model trained on the three databases (BBD, UBD, URBD). Similar to SVR case, the performance of XGBoost is better when it is trained on URBD than on UBD (figures \ref{fig:comp2}(a), \ref{fig:comp2}(c)). In panel \ref{fig:comp2}(b), we compare the performance of XGBoost trained on BBD vs URBD. Clearly, when very high accuracy is required, the XGBoost trained on much fewer data points collected via BO (BBD) surpass the accuracy of the same model trained on uniform or a combination of uniform and random data generation. The same conclusion can be reached by calculating the MSE instead of $R^2$ (figure \ref{fig:comp2}(c)).

We see from figure \ref{fig:comp2}(b), if a very high accuracy of $R^2  \approx 0.97$ for the ML prediction is required, this can be achieved with only 77 data points obtained by BO approach, in our complex 6D parameter space (figure \ref{fig:comp2}b). However, ML model trained on the other two databases could not reach this accuracy, for all the data points that we included in the training (729 data points). This supremacy implemented to 3D FDTD simulations within this study, means that the XGBoost can reach a very high predictive accuracy of $R^2  \approx 0.97$, trained on a database generated with 97944 core-hour of simulation. If this needs to be achieved with more traditional way of generating database, it requires an order of magnitude longer simulation time (927288 core-hour of simulations). This difference becomes more apparent with even more time consuming simulations or experiments or in a more complex parameter space. We can see this from figure\ref{fig:comp2}(d). 

Similar to the 6D parameter space that we mentioned thus far, we repeat our procedure for two other dimensions of parameter space (4D and 5D). For each case, we build three databases (UBD, URBD, BBD) as explained before and we compare the performance of optimized XGBoost trained on each database. Figure \ref{fig:comp2}(d) shows the relative efficiency as a function of dimension of parameter space. The relative efficiency here is obtained by dividing the number of data points required for XGBoost trained on UBD (blue diamonds) or URBD (orange circles) to reach $R^2 \approx 0.97$ by the number of data points for BBD to surpass this accuracy. As the parameter space becomes more complex (its dimension increases) the gap between the models trained on BBD vs the other methods grow wider. In 6D parameter space one requires an order of magnitude larger database obtained using uniform distribution of data points than the one obtained using Bayesian approach. We expect that a further increase in the dimensionality of the parameter space or its complexity, will significantly widen the gap between the Bayesian optimization-driven approach and the other two. This is due to the fact that, uniform sampling of each parameter is not optimal, contrary to BO, where less important parameter sets are sampled sporadically. 

The other benefits of BO driven approach is the possibility to smoothly increase the number of data points to reach required predictive precision for ML model. On the other hand in the UBD the ML accuracy does not increase steadily by adding each data point. The large increase in the accuracy of ML happens when the size of the database is equal to the evenly divisible numbers. If the dimension of the parameter space increases, the gaps between evenly divisible points increases, thus makes BO approach even more important.

It is worth noting that, this methodology can be applied to less time-consuming models (eg., 2D FDTD simulations in this case), to yield more informative data points. Then these data points can be directly used for more accurate but time-consuming simulations (eg., 3D FDTD simulations). The broader implications of our research extend beyond the current problem of predicting reflectance spectra from Bragg grating characteristics in photonics. It has the potential to impact data-driven modeling in various scientific and engineering domains. By enabling the collection of the minimal amount of data necessary for high-accuracy predictions. Our approach promises cost savings, accelerated model development, and increased practicality in data-driven modeling.

\section{Conclusion} 
This research represents a step toward addressing the challenge of efficient construction of a database for training ML models that balances minimal data collection with its high degree of informative quality. 
To accomplish this goal, we employed BO with an acquisition function, designed to emphasize exploration over exploitation (including only the standard deviation). By doing this, we prioritized the acquisition of diverse and informative data points. We incrementally expanded our database by selecting points that maximize the acquisition function.
We showed that if a lower accuracy is required, one can apply shallow ML models, like SVR. For obtaining higher accuracy, more complicated models, like XGBoost, are more suitable.
Within the range of applicability of each model, the ML method trained on BBD significantly outperformed the same model trained on more traditional data acquisition approaches.
We have demonstrated the effectiveness of BO in this context, shedding light on the complexities of data-driven modeling in high-dimensional parameter spaces and showing the potential for efficient data generation in the training ML models for scientific inquiries. Our findings invite further exploration and application of this methodology across diverse domains, promising to transform the landscape of data-driven research and modeling.\\

\textbf{Author contributions:}
MRM, IN, YR and AW conceived and planned the research. MRM carried out the method development, optimizations and ML simulations. IN implemented the Gaussian process regression in MATLAB for automatic data generation with Lumerical. YR developed the FDTD Bragg grating model. All authors contributed to the interpretation of the results, manuscript writing, and provided feedback.\\

\textbf{Research funding:} This work was supported by the VDI Technologiezentrum GmbH with funds provided by the Federal Ministry of Education and Research under grant no. 13N14906 and the DLR Space Administration with funds provided by the Federal Ministry for Economic Affairs and Climate Action (BMWK) under Grant No 50WK2272.\\

\textbf{Conflict of interest statement:} The authors declare no conflicts of interest regarding this article.\\

\bibliography{References}

\end{document}